\documentclass{article}

\usepackage[preprint, nonatbib]{neurips_2024}

\usepackage[utf8]{inputenc} 
\usepackage[T1]{fontenc}    
\usepackage{hyperref}       
\usepackage{url}            
\usepackage{booktabs}       
\usepackage{amsfonts}       
\usepackage{nicefrac}       
\usepackage{microtype}      
\usepackage[dvipsnames,table,xcdraw]{xcolor}        
\usepackage{graphicx}
\usepackage[numbers,sort]{natbib}
\usepackage{amsmath}
\usepackage{amssymb}
\usepackage{algpseudocode}
\usepackage{algorithm}
\usepackage{multirow}
\usepackage{arydshln}
\usepackage{pifont}
\usepackage{array}
\usepackage{subcaption}
\usepackage[font=small]{caption}
\usepackage{wrapfig}
\usepackage[capitalize]{cleveref}
\usepackage{enumitem}

\hypersetup{
    colorlinks=true,
    linkcolor=red,
    citecolor=RoyalBlue,
    filecolor=magenta,      
    urlcolor=cyan,
}

\definecolor{qingse}{HTML}{177cb0}
\definecolor{juse}{HTML}{ca6924}

\title{Flash-VStream: Memory-Based Real-Time Understanding for Long Video Streams}

\author{%
Haoji Zhang$^{1*}$ \quad Yiqin Wang$^{1*}$ \quad Yansong Tang$^{1\dag}$ \quad Yong Liu$^1$ \\
\textbf{Jifeng Dai}$^2$ \quad \textbf{Jiashi Feng}$^3$ \quad \textbf{Xiaojie Jin}$^{3\dag\ddag}$ \\
$^1$Shenzhen International Graduate School, Tsinghua University \\
$^2$Department of Electronic Engineering, Tsinghua University \quad $^3$ByteDance Inc.\\
\texttt{\{haoji-zh20@mails.,yq-wang23@mails.,tang.yansong@sz.\}tsinghua.edu.cn}\\
\texttt{jinxiaojie@bytedance.com}
}

\begin{document}

\maketitle

\def\thefootnote{$*$}\footnotetext{
Equal contribution. 
\dag Correspondence to Xiaojie Jin <\href{mailto:jinxiaojie@bytedance.com}{jinxiaojie@bytedance.com}> and Yansong Tang <\href{mailto:tang.yansong@sz.tsinghua.edu.cn}{tang.yansong@sz.tsinghua.edu.cn}>.
\ddag Project lead.
Project page \href{https://InvincibleWyq.github.io/vstream-page/}{https://InvinciblWyq.github.io/vstream-page} .
}

\begin{abstract}
Benefiting from the advancements in large language models and cross-modal alignment, existing multi-modal video understanding methods have achieved prominent performance in offline scenario. However, online video streams, as one of the most common media forms in the real world, have seldom received attention. Compared to offline videos, the ``dynamic'' nature of online video streams poses challenges for the direct application of existing models and introduces new problems, such as the storage of extremely long-term information, interaction between continuous visual content and ``asynchronous'' user questions. 
Therefore, in this paper we present Flash-VStream, a video-language model that simulates the memory mechanism of human. Our model is able to process extremely long video streams in real-time and respond to user queries simultaneously. 
Compared to existing models, Flash-VStream achieves significant reductions in inference latency and VRAM consumption, which is intimately related to performing understanding of online streaming video.
In addition, given that existing video understanding benchmarks predominantly concentrate on offline scenario, we propose VStream-QA, a novel question answering benchmark specifically designed for online video streaming understanding. 
Comparisons with popular existing methods on the proposed benchmark demonstrate the superiority of our method for such challenging setting. To verify the generalizability of our approach, we further evaluate it on existing video understanding benchmarks and achieves state-of-the-art performance in offline scenarios as well. All code, models, and datasets are available at the \href{https://invinciblewyq.github.io/vstream-page/}{project page}.
\end{abstract}

\section{Introduction}

Online video streaming is a prevalent media format with a broad spectrum of applications.
In the field of robotics, for instance, robots operating in the wild can leverage stream understanding models
to interpret and react to their environment in real-time~\cite{supancic2017tracking, sermanet2023robovqa}.
Similarly, in surveillance systems, stream understanding models can process and analyze video streams from specific locations continuously,
thereby improving overall security~\cite{chen2019distributed, muhammad2019deepres}. However, best existing large video-language models fails to perform real-time long video question-answering upon user queries~\cite{llamavid,chatunivi,vistallama,moviechat}.
The main reason is that: visual tokens between consecutive frames are heavy and redundant without effective compression, making it
impossible to save all visual features in limited GPU Memory (VRAM), as well as significantly increasing the decoding latency of language model.

Considering how humans process live video streams in real-time can provide inspiration for the design of video stream understanding models.
This procedure can be divided into four steps~\cite{feigenbaum1970information}:
1) Perceiving: human eyes continuously encode an endless visual information into brain.
2) Memorizing: human brain compresses the visual information and update brain memory with it. With limited memory capacity, humans tend to have
clearer detailed memories of recent events while they only remember the most important parts of events from the distant past.
3) Recalling: whenever a person is asked about what happens before, his/her brain retrieve the memory.
4) Answering: human brain integrates the memory information with the context provided by the question, and generate an answer.

\begin{figure}[t]
    \centering
    \includegraphics[width=0.95\linewidth]{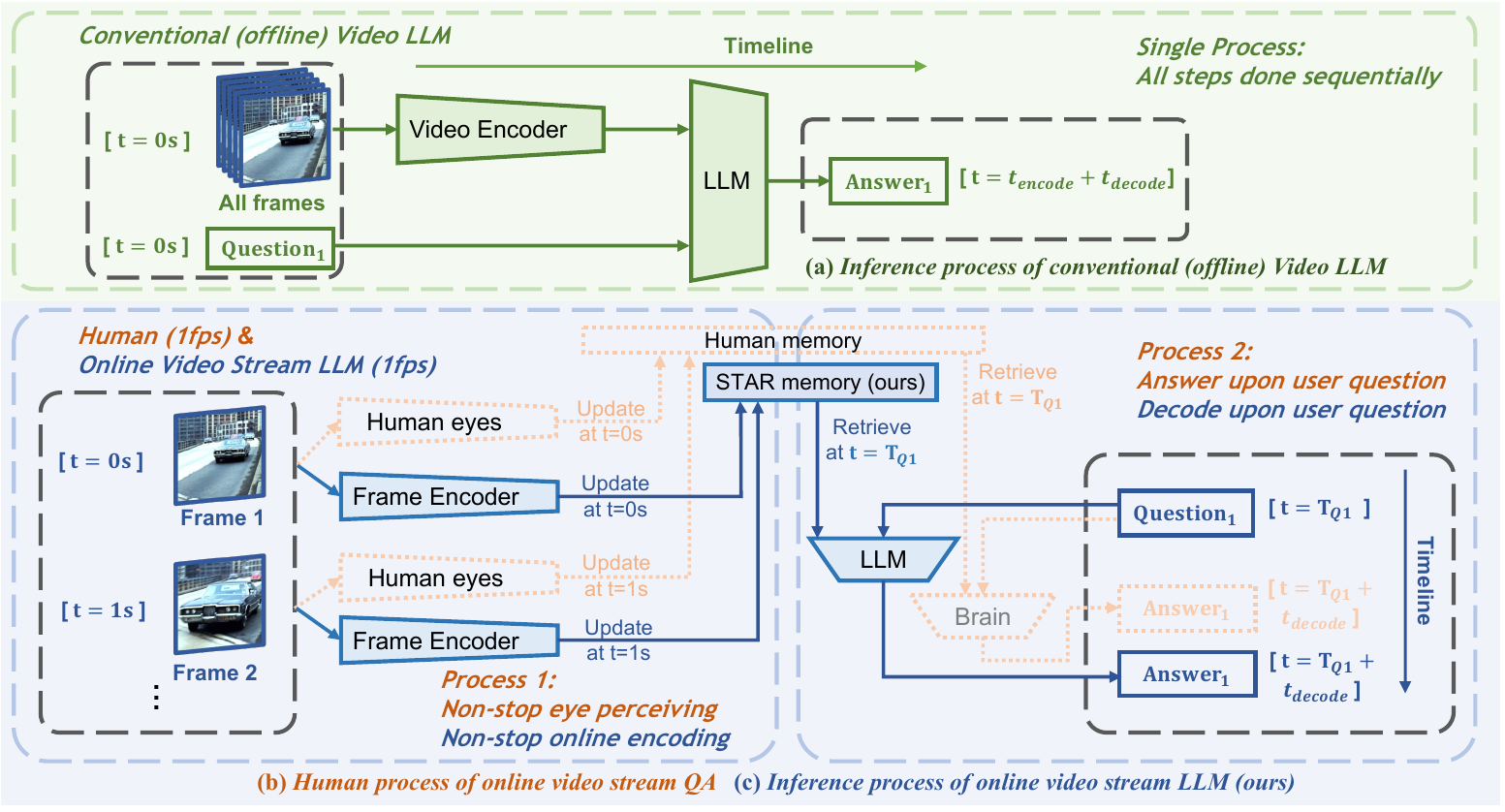}
    \caption{\textbf{Comparing \textcolor{ForestGreen}{(a) conventional offline pipeline} and
    \textcolor{juse}{(b) human processing pipeline} with
    \textcolor{qingse}{(c) our proposed Flash-VStream for online video streaming understanding}.} 
    Zoom in for better view.}
    \label{fig:streaming_model} \vspace{-10pt}
\end{figure}

It is worth noting that the four human processing steps above are not strictly sequential.
As shown in \textcolor{juse}{~\Cref{fig:streaming_model} (b)} (focus on the brown part and ignore the blue part), the first two steps can be performed by a process (on the left),
while the last two steps being performed by another process simultaneously (on the right).
In other words, humans can perceive and memorize new information while recalling and answering questions about the past simultaneously.
While the ``process'' for perceiving and memorizing is always running,
the ``process'' for recalling and answering is only activated upon user questions.
This is the key to online video stream understanding.
In contrast, most existing video-QA methods~\cite{llamavid,chatunivi,vistallama} are based on offline video understanding,
where user query and finite-length video are given to the model at the same time.
As shown in \textcolor{ForestGreen}{~\Cref{fig:streaming_model} (a)},
these methods only consist of the two strictly sequential steps: perceiving and answering.
The lack of a compressed memory mechanism in these offline methods result in a dilemma:
1) If the model keeps the redundant visual tokens of all frames, the high VRAM consumption leads to limited input frame capacity.
2) If the model performs question-aware encoding and only keep those visual tokens that are relevant to the question, it has to re-encode all the
visual information from scratch every time a new query is given, leading to an unacceptable inference latency for online video streams.

To address this challenge, we introduce Flash-VStream, \textit{a video-language model that is able to process extremely long video streams in real-time
    and respond to user queries simultaneously}.
As shown in \textcolor{qingse}{~\Cref{fig:streaming_model} (c)},
Flash-VStream (blue) highly resembles human processing pipeline (brown) in terms of ``4-step, 2-process'' design philosophy.
The frame encoder resembles human eyes and the LLM resembles human brain.
The learnable memory mechanism in Flash-VStream, named \textbf{S}patial-\textbf{T}emporal-\textbf{A}bstract-\textbf{R}etrieved (\textbf{STAR}) memory, is carefully designed to compress
necessary visual information and update memory in a online and real-time manner, as shown in~\Cref{fig:overall_framework}.

In addition, recognizing the limitations of existing offline and short-length video QA benchmarks, for evaluating video stream understanding in online settings,
we propose VStream-QA, a novel question answering benchmark specifically designed for online video stream understanding.
The main features of VStream-QA lies in:
i) Each question-answer pair is marked with a specific timestamp in the video and
only related to the visual information before that timestamp, which is consistent with the online video stream understanding setting.
ii) The video length ranges from 30 minutes to 60 minutes, which is significantly longer than existing benchmarks,
making it capable of evaluating model's performance on extremely long videos.
iii) The videos cover a variety of content, including first-person perspective (ego-centric) videos, and third-person perspective movies.

\begin{figure}[t]
    \centering
    \begin{minipage}{0.46\linewidth}
        \includegraphics[width=1\linewidth]{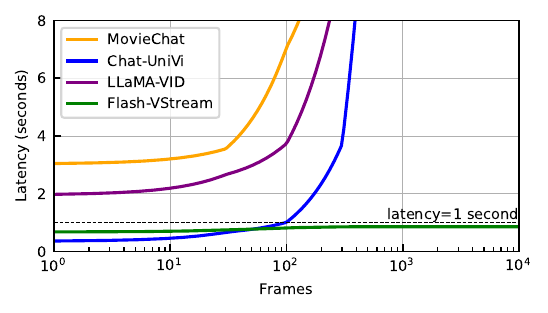} \vspace{-20pt}
        \caption{\textbf{Inference latency (y-axis) v.s. frame number (x-axis).}
            Latency tested on an A100 gpu.
            Our model is able to process extremely long video streams, and
            perform real-time answering within 1 second upon user's query. }
        \label{fig:less_vram_latency}
        \vspace{5pt}
    \end{minipage}
    \hfill
    \begin{minipage}{0.52\linewidth}
        \centering
        \footnotesize
        \setlength{\tabcolsep}{1pt}
        \centering
        \begin{tabular}{lcccccccc}
            \toprule
            \multirow{2}{*}{\textbf{Method}} & \multicolumn{2}{c}{\textbf{RVS-Ego}} & \multicolumn{2}{c}{\textbf{RVS-Movie}} & \multirow{2}{*}{\textbf{VRAM$\downarrow$}} \\
            \cline{2-5}
            \addlinespace[2pt]
            ~ & \textbf{A}  & \textbf{S}  & \textbf{A}  & \textbf{S} \\
            \midrule
            Video-ChatGPT \cite{video-chatgpt}  & 51.0  & 3.7 & 51.7  & 3.3 & \textcolor{gray}{16.62GB *}   \\
            MovieChat     \cite{moviechat}  & 50.7  & 3.4 & 36.0  & 2.3 & 16.90GB $\dagger$                  \\
            Chat-UniVi    \cite{chatunivi} & 51.2  & 3.8 & 51.8  & 3.3 & 77.56GB $\dagger$                  \\
            LLaMA-VID     \cite{llamavid} & 53.4  & 3.9 & 48.6  & 3.3 & 33.64GB $\dagger$                  \\
            \midrule
            \textbf{Flash-VStream} & \textbf{57.3} & \textbf{4.0}& \textbf{53.1} & \textbf{3.3} & \textbf{16.03GB $\dagger$}          \\
            \bottomrule
        \end{tabular}
        \captionof{table}{\textbf{Comparison with SoTA methods on zero-shot real-time VideoQA.} \textbf{A} and \textbf{S} denote accuracy and score, respectively.
        VRAM tested on an A100 gpu.
        *: Tested with a 100-frame input video (maximum support of Video-ChatGPT).
        $\dagger$: Tested with a 1000-frame input video.}
        \label{tab:VStreamQA}
    \end{minipage}
\end{figure}

On these challenging online benchmarks, Flash-VStream achieves state-of-the-art performance,
while achieving significant reductions in inference latency and VRAM consumption as shown in ~\Cref{fig:less_vram_latency} and \Cref{tab:VStreamQA}.
Zero-shot video question answering experiments on 4 conventional offline video QA benchmarks
further prove the generalization ability of Flash-VStream, as shown in ~\Cref{tab:VideoQA}.
Comprehensive ablation studies prove the effectiveness of the memory mechanism we adopted.
We summarize our contribution as follows:

\begin{itemize}[leftmargin=*]
\item We introduce Flash-VStream, a novel large video-language model that is able to process extremely long video streams
      in real-time and respond to user queries simultaneously. A cleverly designed memory mechanism named STAR
      is introduced to compress necessary visual information while leaving out the redundancy between consecutive frames.
\item While maintaining state-of-the-art performance on both online and offline benchmarks,
      Flash-VStream achieves significant reductions in inference latency and GPU Memory (VRAM) consumption,
      enabling online video stream QA in real-time.
\item We also propose VStream-QA, a new QA benchmark specifically designed for video understanding
      in online settings. Its question-answer-timestamp triplet design is consistent with online scenario and
      its video length is significantly longer than existing benchmarks,
      making it capable of evaluating model's performance on nearly-infinite long video streams.
\end{itemize}

\section{Related work}

\textbf{Multi-modal large language models.}
With recent advances in Large Language Models (LLMs)~\cite{brown2020language,Ouyang2022TrainingLM,Touvron2023Llama2O,touvron2023llama}, many works try to build
Multimodal Large Language Models (MLLMs) that integrate text with visual data or other modalities. For instance, the BLIP series~\cite{Li2022BLIPBL,li2023blip,Dai2023InstructBLIPTG}
proposed a efficient strategy for bootstrapping multimodal understanding with pretrained LLMs and image encoders, and the LLaVA series~\cite{llava, Liu2023ImprovedBW}
leverage GPT-generated visual instruction data to tune open language models.
With the development of image-text models, researchers have begun extending image data to videos.
The biggest challenge for Video LLM is how to compress redundant frame features.
LLaMA-VID~\cite{llamavid} represents single-frame features with a few tokens,
Chat-UniVi~\cite{chatunivi} employs dynamic tokens to model image and video features of different scale,
and Vista-LLaMA~\cite{vistallama} uses a sequential visual projector to represent an entire video with fewer tokens.
These methods either requires a multi-step visual encoding process with high latency~\cite{chatunivi},
or have a linearly increasing VRAM cost with the number of frames~\cite{llamavid, vistallama},
making them unsuitable for real-time long video stream understanding.
MovieChat~\cite{moviechat} proposed to combine all frame features through a simple average strategy.
Though it is able to process long video with limited VRAM cost, its performance is suboptimal due to its training-free framework and non-learnable memory mechanism.
In our proposed Flash-VStream, we introduce a learnable memory mechanism that encode frames in a online and real-time manner,
disentangling the visual encoding process and answer decoding process, thus enabling real-time video stream understanding.

\textbf{Real-time video stream understanding.}
Real-time video stream understanding is a challenging task that requires the model to process video streams in real-time and finish specific tasks based on the video.
Most existing real-time methods are designed to perform a single, specific vision task, such as real-time object tracking~\cite{track1, track2, track3} and
real-time action recognition~\cite{action1, action2}. Considering natural language is becoming a general interface for various tasks and modalities~\cite{alayrac2022flamingo, gao2021clip, multimodal3, li2023blip},
our work focuses on real-time video stream question answering upon user queries, which is a more challenging and comprehensive task.

\textbf{Memory mechanism for long sequence processing.}
Memory mechanism is widely used to store and retrieve information in all forms of sequence processing tasks,
such as time series forecasting~\cite{chang2018memory}, recommendation system~\cite{tan2021dynamic}, machine translation~\cite{Daelemans_2005},
and video object segmentation~\cite{cheng2022xmem}.
Inspired by the idea of Neural Turing Machine (NTM)~\cite{NTM}, a learnable mechanism that resembles the working memory system
of human cognition, we proposed a learnable visual memory that is able to compress visual information and update memory
in a online and real-time manner.

\section{Flash-VStream}

\begin{figure}[h]
    \centering
    \includegraphics[width=1.00\linewidth]{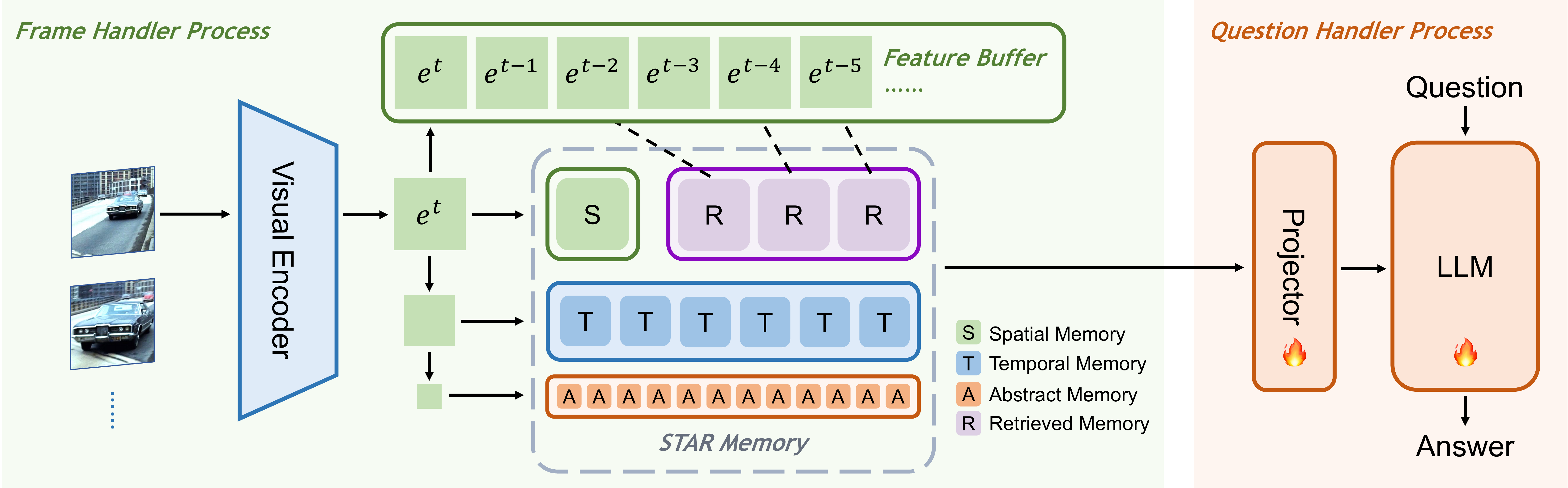} 
    \vspace{-15pt}
    \caption{\textbf{The overview of Flash-VStream framework for real-time online video stream understanding.}
    Flash-VStream is executed by two processes, namely ``frame handle'' and ``question handler''. The frame handler is responsible for encoding frames and writing to memory, which contains a visual encoder, a STAR memory and a feature buffer. The question handler is responsible for reading from memory and answering questions anytime, which contains a projector and a Large Language Model.
    } 
    \label{fig:overall_framework}
\end{figure}

As shown in ~\Cref{fig:overall_framework}, our Flash-VStream framework consists of three main components: 
(1) a streaming visual encoder that continuously processes video frames, 
(2) a \textbf{S}patial-\textbf{T}emporal-\textbf{A}bstract-\textbf{R}etrieved memory mechanism (\textbf{STAR} memory), including memory writing and reading with the help of a feature buffer. 
(3) a LLM decoder capable of providing real-time responses to questions raised by users.
To perform real-time inference, Flash-VStream is deployed in two asynchronous processes. The frame handler process manages the streaming visual encoder and STAR memory consolidation. The question handler process manages the real-time LLM decoder, STAR memory reading and interactions with users. The only connection between these two processes is the shared memory, which can be written by the first process and read by both.

\subsection{Streaming visual encoder}

Like human eyes, the streaming visual encoder can continuously encode visual information into embedded features. We use the pre-trained CLIP ViT-L \cite{radford2021learning} as visual encoder. Only patch tokens are used during training and inference. 
Specifically, given a frame stream $\{V^t\}_{t=1}^{\infty}$, the encoder maps the $t$-th
frame $V^t\in \mathbb{R}^{H\times W\times 3}$ to feature map ${e^t}\in \mathbb{R}^{P\times P\times D}$, where $P\times P$ is the number of ViT patch tokens and $D$ is the hidden dimension of ViT.

\subsection{\underline{S}patial-\underline{T}emporal-\underline{A}bstract-\underline{R}etrieved memory}
\label{sec:memory}

\begin{figure}[h]
    \centering
    \includegraphics[width=1.00\linewidth]{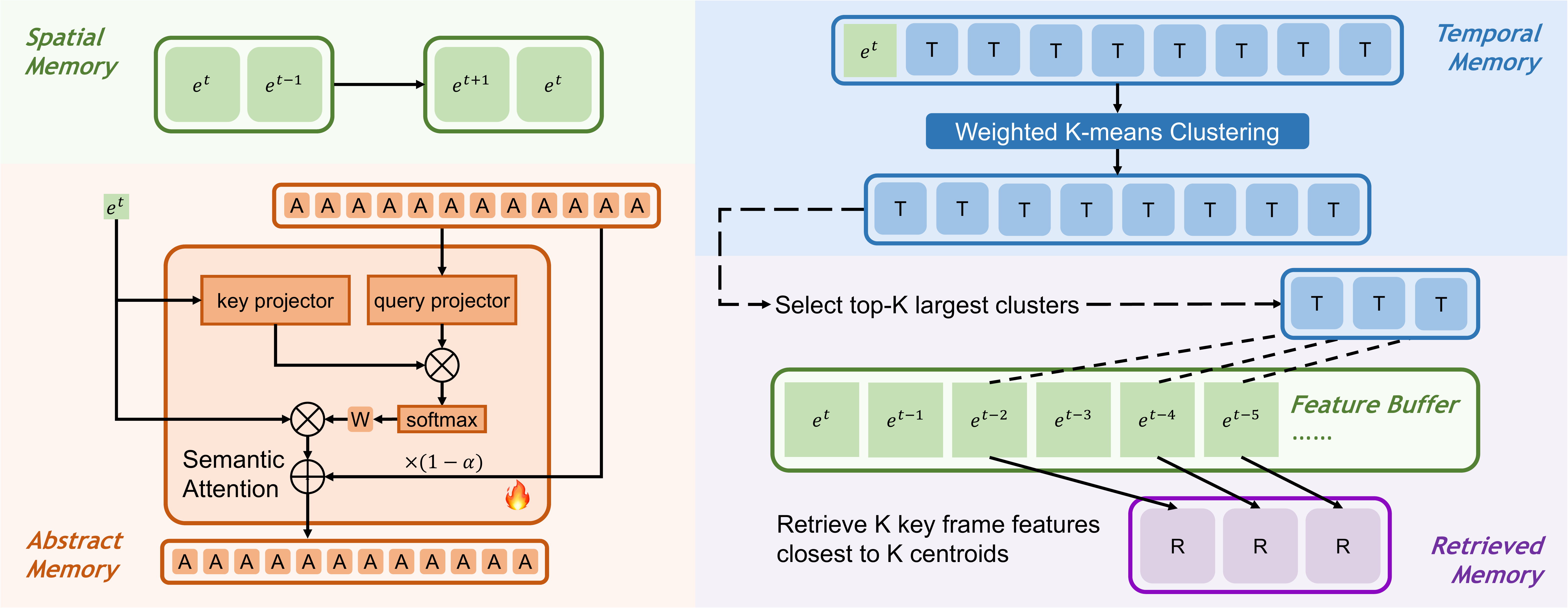} 
    \vspace{-15pt}
    \caption{\textbf{STAR memory writing mechanism.}
    (a) Update spatial memory by a \textit{FIFO} queue. 
    (b) Update temporal memory by \textit{Weighted K-means Clustering}. 
    (c) Update abstract memory by \textit{Semantic Attention}. 
    (d) Update retrieved memory by key frame feature retrival. 
    Here feature map $e^T$ has multiple sizes. ``S'', ``T'', ``A'' and ``R'' represent tokens of spatial, temporal, abstract and retrieved memory, respectively.
    } 
    \vspace{-10pt}
    \label{fig:memory_mechanism}
\end{figure}

In order to handle information of different levels of granularity, we design a STAR memory with 4 components: 
spatial memory $M_{\text{spa}} \in \mathbb{R}^{N_{\text{spa}} \times P_{\text{spa}}^2 \times D}$, 
temporal memory $M_{\text{tem}} \in \mathbb{R}^{N_{\text{tem}}\times P_{\text{tem}}^2 \times D}$, 
abstract memory $M_{\text{abs}} \in \mathbb{R}^{N_{\text{abs}}\times P_{\text{abs}}^2 \times D}$ and 
retrieved memory $M_{\text{ret}} \in \mathbb{R}^{N_{\text{ret}}\times P_{\text{spa}}^2 \times D}$. 
A feature buffer $M_{\text{buff}} \in \mathbb{R}^{N_{\text{buff}}\times P_{\text{spa}}^2 \times D}$ is used to store the feature of latest $N_{\text{buff}}$ frames. 
Therefore, the overall memory size is limited to $\text{MAXSIZE}=(N_{\text{spa}}+N_{\text{ret}})\times P_{\text{spa}}^2 + N_{\text{tem}}\times P_{\text{tem}}^2 + N_{\text{abs}}\times P_{\text{abs}}^2$ tokens.

\textbf{Spatial memory.} Spatial memory houses the most recent and detailed spatial information for short-term use, implemented as a \textit{FIFO} (First-In-First-Out) queue, as illustrated in \Cref{fig:memory_mechanism} and \Cref{equ:spa}. This architecture enables continuous updating with the newest frames, facilitating immediate access to fine-grained spatial data.

\textbf{Temporal memory.} Temporal memory integrates dynamic information over time, crucial for long-term retention. When its size surpasses $N_{\text{tem}}$, the $g_{\text{wkmeans}}$ (\textit{Weighted K-means Clustering}) algorithm is applied, as shown in \Cref{equ:tem} and \Cref{alg:wkmeans}. This strategy condenses the memory content into $N_{\text{tem}}$ clusters which can be seen as the representation of key events in videos. Then the centroids of these clusters are used as the new memory for efficiently storing temporal contexts.

\textbf{Abstract memory.} Abstract memory supports high-level semantic concept interpretation through $f_{SA}$, the \textit{Semantic Attention} model. It follows \Cref{equ:abs} to synthesize the insights gained from both spatial and temporal memories into abstracted, actionable knowledge. $f_{SA}$ keeps adjusting $M_{\text{abs}}$, the synopsis of whole video by newest features.
Refer to \Cref{fig:memory_mechanism} and \Cref{alg:sa} for details.

\textbf{Retrieved memory.} Retrieved memory focuses on recalling precise spatial details by identifying and retrieving the most substantial frame features. As shown in \Cref{fig:memory_mechanism}, it first selects the  top-K (where K equals $N_{\text{ret}}$) largest clusters from the $N_{\text{tem}}$ clusters obtained in temporal memory $M_{\text{tem}}$. Then the nearest frame features in feature buffer to centroids of these K clusters are retrieved to supplement the temporal memory with more detailed spatial information. This process is illustrated in \Cref{equ:ret} and \Cref{alg:ret}.

In brief, a new feature $e^t$ is written to STAR memory as follows: 
\begin{align}
    M_{\text{buff}}^t &= \texttt{concat}\big(g_{\text{pooling}}(e^t, P_{\text{spa}}) , M_{\text{buff}}^{t-1}\big) [0:N_{\text{buff}},:,:] \label{equ:buff} \\
    M_{\text{spa}}^t &= M_{\text{buff}}^t[0:N_{\text{spa}},:,:] \label{equ:spa} \\
    M_{\text{tem}}^t &= g_{\text{wkmeans}}\Big( 
    \texttt{concat}\big(g_{\text{pooling}}(e^t, P_{\text{tem}}) , M_{\text{tem}}^{t-1}\big), N_{\text{tem}}\Big) \label{equ:tem} \\
    M_{\text{abs}}^t &= f_{SA}\big(M_{\text{abs}}^{t-1}, g_{\text{pooling}}(e^t, P_{\text{abs}}), N_{\text{abs}}\big) \label{equ:abs} \\
    M_{\text{ret}}^t &= g_{\text{retrieve}}(M_{\text{buff}}^t, M_{\text{tem}}^t, N_{\text{ret}}) \label{equ:ret}
\end{align}
\vspace{-10pt}

Here $g_{\text{pooling}}(e,P^\prime)$ applies \textit{Average Pooling} to compress feature map $e$ from $P^{2}$ to $P^{\prime 2}$ size along width and height dimensions. $\texttt{concat}(a,b)$ means concatenating tensors $a$ and $b$ along time axis.

\subsection{Real-time LLM decoder}
The LLM decoder works as part of a real-time question answering server. When triggered by a question $Q^t$ at time $t$, the LLM decoder first calculates the text embedding $I_{\text{text}}^t = f_{\text{embed}}(Q^t)$ and maps the STAR memory $M^t=M_{\text{spa}}^t+M_{\text{tem}}^t+M_{\text{abs}}^t+M_{\text{ret}}^t$ to embedding space with the projector $I_{\text{vision}}^t = f_{\text{proj}}(M^t)$. Then it starts to generate answer $A^t = f_{\text{LLM}}(I_{\text{text}}^t, I_{\text{vision}}^t).\text{decode}()$ in real time.

\subsection{Implementation details}
In this study, we utilize pre-trained CLIP~ViT-L/14-224px~\cite{radford2021learning} as streaming visual encoder.
Following LLaVA~\cite{liu2023visual}, we choose a 2-layer-MLP as visual projector and pre-trained Vicuna-7B~\cite{vicuna2023} as LLM decoder. 
Considering the balance between performance and resource consumption, we set $P_{\text{spa}}=8$, $P_{\text{tem}}=4$, $P_{\text{abs}}=1$, $N_{\text{buff}}=300$, $N_{\text{spa}}=1$, $N_{\text{tem}}=N_{\text{abs}}=25$ and $N_{\text{ret}}=3$. The MAXSIZE of STAR memory is set to 681 tokens in order to keep computational efficiency.

We train Flash-VStream for 2 stages: modality alignment and instruction tuning. The training data keep the same with LLaMA-VID~\cite{llamavid}, including LLaVA-filtered-558K \cite{llava} image-caption pairs and LLaMA-VID-filtered-232K \cite{llamavid} video-caption pairs for stage 1, LLaVA-filtered-665K \cite{llava} image QA pairs and Video-ChatGPT-filtered-98K \cite{video-chatgpt} video QA pairs for stage 2. For each stage, the model is trained for 1 epoch on 8 A100 80G GPUs. During training, the parameters of visual encoder are frozen and the parameters of LLM are frozen only for the first stage. All training and inference experiments was conducted under BF16 precision to save time and resources. Other hyper-parameters can be found at Table \ref{tab:train_setting}.

\section{VStream-QA: A new benchmark for online video stream QA}

Previous video QA benchmarks~\cite{xu2017video, xiao2021next, yu2019activitynet} mostly focus on offline video understanding,
where user query and finite-length video are given to the model at the same time.
To our best knowledge, there is no existing benchmark specifically designed for online video stream understanding.
Also, most existing benchmarks are limited to short-length videos within 1 minute~\cite{xu2017video,xiao2021next} or
medium-length videos within 10 minutes~\cite{yu2019activitynet,vistallama,moviechat, egoschema},
which are unsuitable for simulating online video stream.

To address this problem, we propose VStream-QA, a novel question answering benchmark specifically designed for
online video stream understanding. VStream-QA consists of two parts: VStream-QA-Ego and VStream-QA-Movie, which
are designed for evaluating first-perspective ego-centric understanding and third-perspective plot understanding, respectively.
The prominent features of VStream-QA are i) each question-answer pair is marked with a specific timestamp in the video and
only related to the visual information before that timestamp, ii) containing extremely videos (30 minutes to 60 minutes)
that is significantly longer than existing benchmarks, and iii) covering a variety of video sources and question types.

\begin{figure}[ht]
    \begin{minipage}{0.68\linewidth}
        \centering
        \renewcommand\arraystretch{0.9}
        \setlength{\tabcolsep}{5pt}
        \captionof{table}{\textbf{Video QA Benchmark Comparison.} V for video duration, Q for number of questions, and Desc for descriptive.}
        \begin{tabular}{lcccc}
            \toprule
            \textbf{Benchmark}                      & \textbf{Avg V.}                 & \textbf{Total V.}    & \textbf{Q.}           & \textbf{Goal}                                                                 \\
            \midrule
            MSVD-QA~\cite{xu2017video}              & 10s                             & 1.4h                 & 13K                   & Desc. QA                                                                      \\
            MSRVTT-QA~\cite{xu2017video}            & 15s                             & 12.5h                & 73K                   & Desc. QA                                                                      \\
            ActivityNet-QA~\cite{yu2019activitynet} & 112s                            & 25h                  & 8K                    & Desc. QA                                                                      \\
            Next-QA~\cite{xiao2021next}             & 40s                             & 11h                  & 9K                    & Temporal QA                                                                   \\
            CineCLIP-QA~\cite{vistallama}           & 213s                            & 9h                   & 2.5K                  & Movie QA                                                                      \\
            \midrule
            \multirow{2}{*}{\textbf{VStream-QA}}    & \multirow{2}{*}{\textbf{40min}} & \multirow{2}{*}{21h} & \multirow{2}{*}{3.5K} & \multirow{2}{*}{\begin{tabular}{@{}c@{}}Online Video\\Stream QA\end{tabular}} \\
            ~                                       & ~                               & ~                    & ~                     & ~                                                                             \\
            \bottomrule
        \end{tabular}
        \label{fig:benchmark_comparison}
    \end{minipage}
    \hspace{0.1cm}
    \begin{minipage}{0.3\linewidth}
        \centering
        \includegraphics[width=0.95\linewidth]{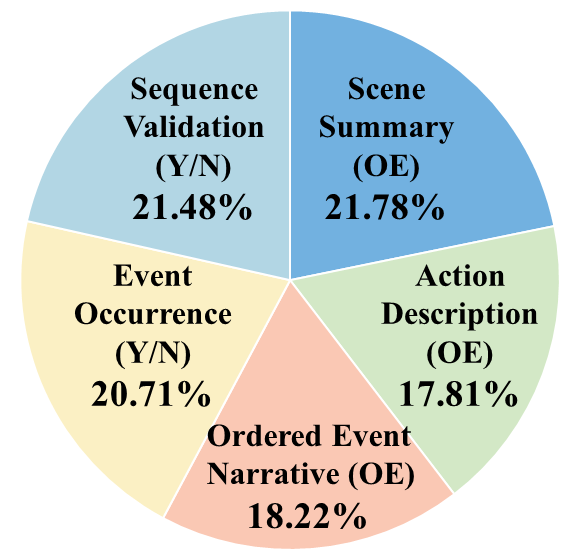}
        \caption{Question Types.}
        \label{fig:question_distribution}
    \end{minipage}
\vspace{-5pt}
\end{figure}

Specifically, VStream-QA-Ego consists of 10 1-hour-long ego-centric video clips from Ego4D dataset~\cite{grauman2022ego4d}
together with 1.5K question-answer-timestamp triplets , while
VStream-QA-Movie consists of 22 half-an-hour-long movie clips from MovieNet dataset~\cite{huang2020movienet}
together with 2K question-answer-timestamp triplets.
As shown in~\Cref{fig:benchmark_comparison}, these two parts consist of a total of 21 hours of video and 3.5K question-answer pairs.
Our proposed VStream-QA fills the gap in existing benchmarks for online video stream understanding,
and provides a extremely long video test set that can be used to evaluate in
both online settings and conventional offline settings.

We carefully design 5 types of questions to evaluate the model's ability to understand both scene content and temporal information.
As shown in~\Cref{fig:question_distribution}, the question types are well balanced.
Specifically, [Scene Summary] and [Action Description] are open-ended questions designed to evaluate the model's ability to understand static and dynamic scene content.
    [Event Occurrence] are yes/no questions designed to evaluate the model's ability to detect whether a specific event or scene occurs in the video.
    [Ordered Event Narrative] and [Sequence Validation] are both designed to evaluate the model's ability to understand the temporal order of events in the video,
with the former being open-ended and the latter being yes/no questions. 
For yes/no questions, its answer ratio is well balanced with 46.3\% yes and 53.7\% no.

In order to balance the annotation quality, 
the data scale, and the total annotation expenses,
we designed a 5-steps data generation pipeline as follows:
1) Video Selection; 2) Dense Captioning; 3) Summary Generation;
4) Question-Answer Generation; and 5) Human Filtering.
For details of each steps, please refer to \Cref{sec:data_gen}.

\section{Experiment}

\subsection{Experimental setup}

\textbf{Datasets.}
For the purpose of real-time video stream understanding, it is crucial for models to keep accurate and efficient. 
To evaluate real-time understanding ability and computational efficiency of models, we them models on Realtime-VStream-QA-Ego/Movie datasets (or RVS-Ego/Movie for short). The real-time version of VStream-QA differentiates normal version by ensuring each question grounded before a predefined timestamp.
To evaluate the basic question answering capability of Flash-VStream, we conduct zero-shot open-ended video question answering experiments on ActivityNet-QA \cite{yu2019activitynet}, NExT-QA \cite{xiao2021next}, MSVD-QA \cite{xu2017video}, MSRVTT-QA \cite{xu2017video} and the proposed VStream-QA-Ego/Movie datasets (or VS-Ego/Movie for short).

\textbf{Evaluation Metrics.} 
For open-ended video question answering tasks, we adopt GPT-3.5 metric following common practices in~\cite{yang2022frozenbilm, li2023videochat, zhang2023llama, zhang2023videollama, video-chatgpt, luo2023valley, moviechat, llamavid, vistallama, chatunivi, lin2023videollava}. With question, ground truth answer and the prediction generated by model, GPT-3.5 is able to judge whether this prediction is correct and provide a score between 0 and 5. We report the GPT-3.5 accuracy and score of each model on VQA datasets.
For computational efficiency test, we report the average respond latency (from questioning to answering) and maximum video random-access memory (VRAM) of models.

\subsection{Zero-shot video question answering}

\begin{table}
    \caption{\textbf{Comparison with SoTA methods on zero-shot VideoQA.}
    \textbf{Acc.} and \textbf{Sco.} denote accuracy and score, respectively.
    *: Evaluated by us. 
    }
    \label{tab:VideoQA}
    \centering
    \fontsize{8.3pt}{9pt}\selectfont
    \setlength{\tabcolsep}{5.5pt}
    \begin{tabular}{lcccccccccccc}
        \toprule
        \multirow{2}{*}{\textbf{Method}} & \multicolumn{2}{c}{\textbf{ActNet}} & \multicolumn{2}{c}{\textbf{NExT}} & \multicolumn{2}{c}{\textbf{MSVD}} & \multicolumn{2}{c}{\textbf{MSRVTT}} & \multicolumn{2}{c}{\textbf{VS-Ego}} & \multicolumn{2}{c}{\textbf{VS-Movie}}                                                                                              \\
        \cline{2-13}
        \addlinespace[1.5pt]
        ~ & \textbf{Acc.} & \textbf{Sco.} & \textbf{Acc.} & \textbf{Sco.} & \textbf{Acc.} & \textbf{Sco.} & \textbf{Acc.} & \textbf{Sco.} & \textbf{Acc.} & \textbf{Sco.} & \textbf{Acc.} & \textbf{Sco.}       \\
        \midrule
        Video-ChatGPT \cite{video-chatgpt}  & 35.2 & 2.7& 54.6 & 3.2  & 64.9 & 3.3 & 49.3& 2.8& 51.7& 3.7& 54.4& 3.4          \\
        MovieChat     \cite{moviechat}      & 45.7 & 3.4& 49.9 & 2.7  & 75.2 & 3.8 & 52.7& 2.6& 52.2& 3.4& 39.1& 2.3          \\
        Chat-UniVi    \cite{chatunivi}      & 45.8 & 3.2& 60.8*& 3.3  & 65.0 & 3.6 & 54.6& 3.1& 50.9& 3.8& 54.0& 3.4          \\
        Vista-LLaMA   \cite{vistallama}     & 48.3 & 3.3& 60.7 & 3.4  & 65.3 & 3.6 & 60.5& 3.3& -   & -  & -   & -            \\
        LLaMA-VID     \cite{llamavid}       & 47.4 & 3.3& 60.3*& 3.4  & 69.7 & 3.7 & 57.7& 3.2& 54.8& 3.9& 51.4& 3.4          \\
        \midrule
        \textbf{Flash-VStream} & \textbf{51.9}& \textbf{3.4} & \textbf{61.6}& \textbf{3.4} & \textbf{80.3}& \textbf{3.9}& \textbf{72.4} & \textbf{3.4} & \textbf{59.0} & \textbf{3.9} & \textbf{56.1} & \textbf{3.4} \\
        \bottomrule
    \end{tabular}
\end{table}

As our model is only trained on \cite{bain2021frozen, huang2020movienet, llava, video-chatgpt}, we compare Flash-VStream with other competitive methods Video-ChatGPT\cite{video-chatgpt}, MovieChat\cite{moviechat}, Chat-UniVi\cite{chatunivi}, Vista-LLaMA\cite{vistallama} and LLaMA-VID\cite{llamavid} on zero-shot real-time VideoQA datasets in Table \ref{tab:VStreamQA}, and on normal zero-shot VideoQA datasets in Table \ref{tab:VideoQA}.
Video-ChatGPT uses temporal pooling and spatial pooling for video understanding. This simple method performs well in real-time movie understanding. 
MovieChat implements a merge-based memory consolidation and uses a Q-Former \cite{Li2022BLIPBL} as feature aggregator. Although it is competitive in understanding some short-video scenes, it falls behind in the domain of extremely long-video understanding, such as with RVS-Ego and RVS-Movie, as shown in \Cref{tab:VStreamQA}.
The newly proposed Chat-UniVi and LLaMA-VID have relative high performances on real-time video understanding benchmark. However, the high computation burden and high latency make it difficult to deploy them for real-time understanding scenes. Flash-VStream achieves SoTA on these benchmarks, demonstrating the proposed STAR memory's exceptional capabilities in information compression and long video comprehension.

\subsection{Computational efficiency}

We measure the inference latency of each model by counting the respond wall time of the question handler process, as presented in \Cref{fig:less_vram_latency}. 
For many models, the inference latency scales up with number of frames because their architectures demand processing all frames at once.
Distinct from them, Flash-VStream leverages an efficient multiprocessing STAR memory mechanism (see \Cref{sec:memory}) for streaming processing frames, which allows relative low inference latency and VRAM cost (detailed in \Cref{tab:VStreamQA}). These attributes enable real-time inference.

\subsection{Ablation study}

\paragraph{Effect of components of memory mechanism.}
\begin{wrapfigure}{r}{6.5cm}
\centering
\footnotesize
\vspace{-10pt}
\renewcommand\arraystretch{0.7}
\captionof{table}{Ablation studies of STAR memory}
\label{ab:star}
\vspace{-5pt}
\begin{tabular}{*{4}{>{\centering\arraybackslash}p{0.3cm}}*{4}{>{\centering\arraybackslash}p{0.4cm}}}
            \toprule
            \multicolumn{4}{c}{\textbf{Memory Type}} & \multicolumn{2}{c}{\textbf{VS-ego}} & \multicolumn{2}{c}{\textbf{VS-movie}} \\ 
            \cline{5-8}
            \addlinespace[3pt]
            S & T & A & R & \textbf{A} & \textbf{S} & \textbf{A} & \textbf{S}  \\
            \midrule
            {\scalebox{0.75}{\color{red}\ding{55}}} & {\scalebox{0.75}{\color{green}\ding{51}}} & {\scalebox{0.75}{\color{green}\ding{51}}} & {\scalebox{0.75}{\color{green}\ding{51}}} & 57.3 & 3.9 & 54.2 & 3.4 \\ 
            
            {\scalebox{0.75}{\color{green}\ding{51}}} &  {\scalebox{0.75}{\color{red}\ding{55}}}  & {\scalebox{0.75}{\color{green}\ding{51}}} & {\scalebox{0.75}{\color{red}\ding{55}}} & 55.1 & 3.9 & 51.4 & 3.4 \\ 
            
            {\scalebox{0.75}{\color{green}\ding{51}}} & {\scalebox{0.75}{\color{green}\ding{51}}} & {\scalebox{0.75}{\color{red}\ding{55}}} & {\scalebox{0.75}{\color{green}\ding{51}}} & 57.0 & 4.0 & 54.1 & 3.4 \\ 
            
            {\scalebox{0.75}{\color{green}\ding{51}}} & {\scalebox{0.75}{\color{green}\ding{51}}} & {\scalebox{0.75}{\color{green}\ding{51}}} & {\scalebox{0.75}{\color{red}\ding{55}}} & 58.0 & 3.9 & 54.4 & 3.4 \\ 
            
            {\scalebox{0.75}{\color{green}\ding{51}}} & {\scalebox{0.75}{\color{green}\ding{51}}} & {\scalebox{0.75}{\color{green}\ding{51}}} & {\scalebox{0.75}{\color{green}\ding{51}}} & \textbf{59.0} & 3.9 & \textbf{56.1} & 3.4 \\ 
            \bottomrule
        \end{tabular}
        \vspace{-5pt}
\end{wrapfigure}
We conduct an ablation study to evaluate the effects of key components of the STAR memory mechanism, i.e., spatial, temporal, abstract and retrieved memory. 
Removing temporal memory can cause a severe performance drop (as shown in the second row of \Cref{ab:star}), indicating that temporal memory is vital in long video stream understanding, as it enables the integration of contextual information across frames for coherent comprehension.
Other types of memory also contribute a lot as they capture different aspect of visual information, such as spatial layout, high-level concepts and pivotal experiences.

\paragraph{Semantic Attention.}
\begin{wrapfigure}{r}{7.5cm}
\centering
\footnotesize
\vspace{-10pt}
\renewcommand\arraystretch{0.7}
\captionof{table}{Semantic Attention v.s. other updating strategies}
\label{ab:semantic_attention}
\begin{tabular}{ccccc}
            \toprule
            \multirow{2}{*}{\textbf{Abstract memory}} & \multicolumn{2}{c}{\textbf{VS-ego}} & \multicolumn{2}{c}{\textbf{VS-movie}} \\ 
            \cline{2-5}
            \addlinespace[3pt]
            ~& \textbf{A} & \textbf{S} & \textbf{A} & \textbf{S}  \\
            \midrule
            Q-Former            & 57.1 & 3.9 & 50.4 & 3.3 \\ 
            Sequential Q-Former & 56.0 & 3.9 & 51.4 & 3.3 \\ 
            Semantic Attention & \textbf{59.0} & \textbf{3.9} & \textbf{56.1} & \textbf{3.4} \\ 
            \bottomrule
        \end{tabular}
\end{wrapfigure}
We compare the proposed Semantic Attention with other memory updating 
strategies as shown in \Cref{ab:semantic_attention}. 
Q-Former~\cite{li2023blip} is widely used by many models~\cite{moviechat, llamavid, zhang2023videollama} and Sequential Q-Former is used by~\cite{vistallama}.
These updating methods are all transformer-based. 
Despite its lightweight nature, the Semantic Attention model outperforms other methods by a large margin. We suppose the reason is that the training dataset is too small for Q-Former based model to adequately learn. The architecture of Semantic Attention facilitates the extraction of key information and the selectively forgetting of irrelevant details, enhancing the model's ability to comprehend abstract concepts in long videos.

\begin{table*}
    \caption{\textbf{Comparison of different spatial and temporal size of STAR memory.}
    \textbf{A} and \textbf{S} denote accuracy and score, respectively.}
    \label{ab:size}
    \vspace{-5pt}
    \footnotesize
    \begin{subtable}[t]{.55\linewidth}
        \centering
        \caption{Spatial Size}
        \label{ab:spatial_size}
        \renewcommand\arraystretch{0.4}
        \setlength{\tabcolsep}{5pt}
        \begin{tabular}{cccccccc}
            \toprule
            \multicolumn{4}{c}{\textbf{Method}} & \multicolumn{2}{c}{\textbf{VS-ego}} & \multicolumn{2}{c}{\textbf{VS-movie}} \\ 
            \cline{5-8}
            \addlinespace[3pt]
            $P_{\text{spa}}$ & $P_{\text{tem}}$ & $P_{\text{abs}}$ & $P_{\text{spa}}$ & \textbf{A} & \textbf{S} & \textbf{A} & \textbf{S}  \\
            \midrule
            16 & 4 & 1 & 16 & 55.7 & 3.9 & 52.1 & 3.3 \\ 
            4  & 4 & 1 & 4 & 58.4 & 4.0 & 53.1 & 3.4 \\ 
            8  & 8 & 1 & 8 & 58.2 & 3.9 & 53.4 & 3.4 \\ 
            8  & 1 & 1 & 8 & 56.4 & 4.0 & 51.9 & 3.4 \\ 
            8  & 4 & 4 & 8 & 57.2 & 4.0 & 54.8 & 3.5 \\ 
            8  & 4 & 1 & 8 & \textbf{59.0} & 3.9 & \textbf{56.1} & 3.4 \\ 
            \bottomrule
        \end{tabular}
    \end{subtable}%
    \begin{subtable}{.35\linewidth}
        \centering
        \caption{Temporal Length}
        \label{ab:temporal_length}
        \setlength{\tabcolsep}{3pt}
        \begin{tabular}{cccccccc}
        \toprule
        \multicolumn{4}{c}{\textbf{Method}} & \multicolumn{2}{c}{\textbf{VS-ego}} & \multicolumn{2}{c}{\textbf{VS-movie}} \\ 
        \cline{5-8}
        \addlinespace[3pt]
        $N_{\text{spa}}$ & $N_{\text{tem}}$ & $N_{\text{abs}}$ & $N_{\text{ret}}$ & \textbf{A} & \textbf{S} & \textbf{A} & \textbf{S}  \\
        \midrule
        1 & 32 & 32 & 3 & 55.3 & 3.9 & 54.9 & 3.4 \\ 
        1 & 16 & 16 & 3 & 57.2 & 3.9 & 54.2 & 3.4 \\ 
        1 & 8  & 8  & 3 & 56.7 & 3.9 & 53.2 & 3.4 \\ 
        1 & 25 & 25 & 3 & \textbf{59.0} & 3.9 & \textbf{56.1} & 3.4 \\ 
        \bottomrule
    \end{tabular}
    \end{subtable}
    \vspace{-10pt}
\end{table*}

\paragraph{Design of spatial size and temporal length of memory.} 
In \Cref{ab:size}, we evaluate how spatial size and temporal length of memory influence long video understanding tasks. For spatial size of memory, although a smaller feature map is harmful to the performance, an excessively larger feature map is not an optimal choice either (see the first row of \Cref{ab:spatial_size}). 
A similar pattern can be observed by varying temporal length of memory in \Cref{ab:temporal_length}, in line with findings from \cite{xu2024pllava}. 
Considering the expensive computational cost of larger and longer memory, we adopt a balanced design.

\subsection{Memory token visualization}

\begin{figure}[h]
    \centering
    \includegraphics[width=\linewidth]{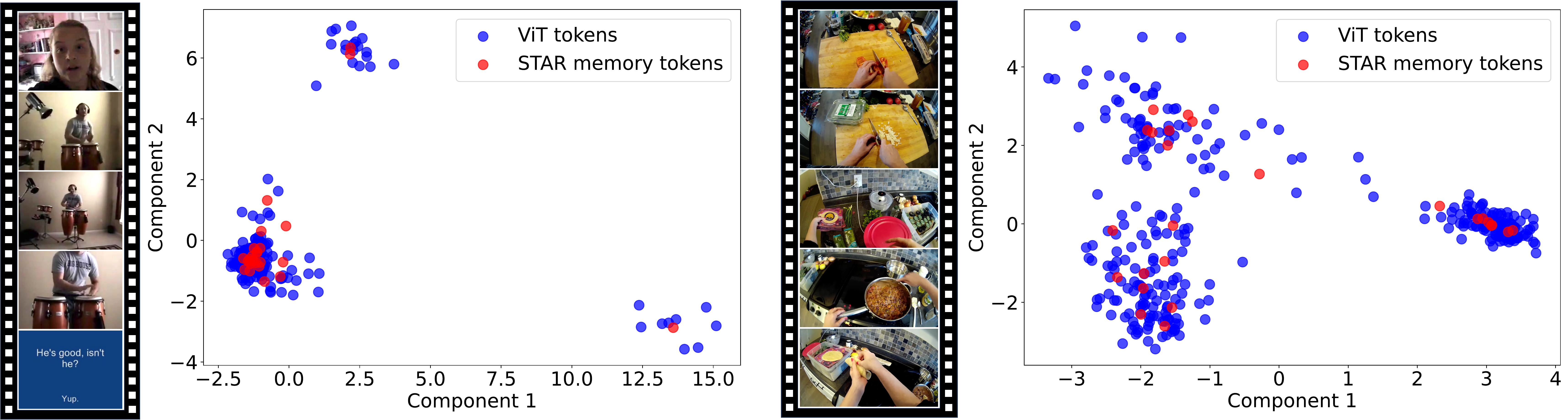} 
    \caption{\textbf{PCA Visualization of memory tokens.}
    Red points represent memory tokens and blue points represent raw vision tokens from visual encoder.
    Left: an example from ActivityNet. Right: an example from Ego4D.
    } 
    \vspace{-10pt}
    \label{fig:memory_visualization}
\end{figure}

\begin{figure}[h]
    \centering
    \includegraphics[width=0.85\linewidth]{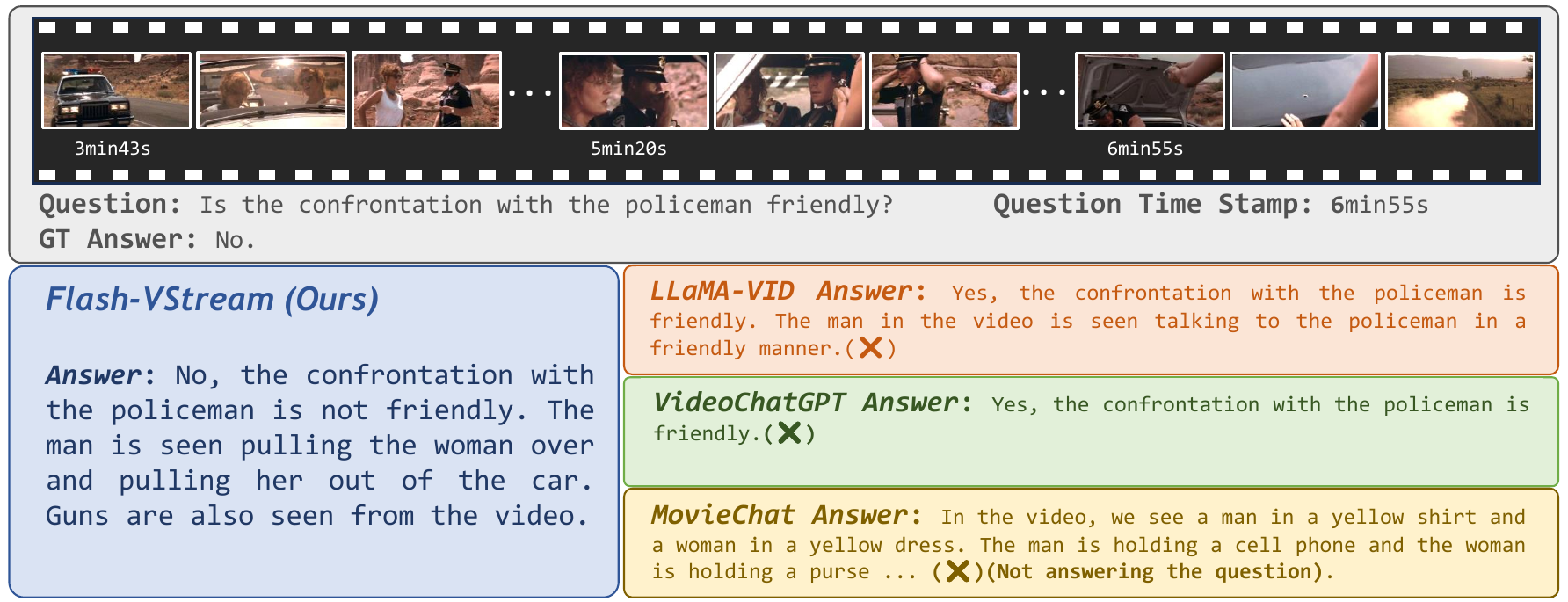} 
    \caption{\textbf{Comparison of different video LLMs on VStream-QA-Movie.} Zoom in for a better view. In this video, a policeman pulls over a vehicle driven by a couple, but they point a gun at him and kill him. Our \textcolor{qingse}{Flash-VStream} is the only model that successfully understands the theme of this long movie clip.}
    \label{fig:case_study} 
    \vspace{-5pt}
\end{figure}

We investigate the memory consolidation procedure in deep feature space.
Specifically, in the left part of~\Cref{fig:memory_visualization}, when inputting a video stream containing 3 significantly different scenes (talking, playing the drums and end credits), the memory will focus on the scene with the longest duration, just like what human will do in their minds.
Relatively static scenes and relatively dynamic scenes are both given lots of attention, as shown in the right part of~\Cref{fig:memory_visualization}. 
The visualization proves that memory tokens effectively reveal the distribution of the vision tokens.

\subsection{Case study}

To better demonstrate the feature of VStream-QA as well as the
effectiveness of Flash-VStream model, we hereby
provide a case study on VStream-QA-Movie dataset.
As shown in ~\Cref{fig:case_study},
a question timestamp is equipped with each question-answer pair, indicating the time when the question is asked.
Models are only provided with the visual content before the question timestamp.
Thanks to the carefully designed STAR memory mechanism,
our \textcolor{qingse}{Flash-VStream} grasp the key visual information and
turns out to be the only model that successfully understands the theme of this long movie clip,
while LLaMA-VID, VideoChatGPT and VStream-QA fail to do so for various reasons.
This proves the effectiveness of our proposed Flash-VStream model in long video understanding tasks.
Refer to model generated answers and the figure caption for details.

\section{Conclusion}

In conclusion, we have introduced Flash-VStream, a video-language model for real-time processing of online video streams and answering user questions. It incorporates a smartly designed memory called STAR, and significantly reduces inference latency and VRAM consumption. In addition, we have proposed a new benchmark for online video understanding called VStream-QA. Our model outperforms existing methods on this new online benchmark and maintains SoTA performance on offline video understanding benchmarks. We hope our work could inspire further research and advancements in the field of online video stream understanding.

{
\bibliographystyle{splncs04}
\bibliography{main}
}
\clearpage


\appendix

\section*{Appendix}
\section{Memory implementation details}

            

This section describes the details of the proposed Spatial-Temporal-Abstract-Retrieved memory mechanism in \Cref{sec:memory}. The STAR memory has both parametric and non-parametric updating strategies. Spatial memory uses simple replacing method. 

As shown in \Cref{alg:wkmeans}, temporal memory performs a \textit{Weighted K-means Clustering Algorithm} temporal-wise to condense $(N_{\text{tem}}+1) \times P_{\text{tem}}^2$ tokens to $N_{\text{tem}} \times P_{\text{tem}}^2$ tokens. Each frame feature in temporal memory $M_{tem}^{(i)}=c_{i} \in \mathbb{R}^{P_{\text{tem}}^2}$ represents the centroid of the i-th feature cluster.

\begin{algorithm}
    \footnotesize
    \caption{Weighted K-means Clustering Algorithm}
    \label{alg:wkmeans}
    \begin{algorithmic}[1] 
        \Require Current temporal memory $\mathbf{M_{\text{tem}}} = \{M_{\text{tem}}^1, M_{\text{tem}}^2, \dots, M_{\text{tem}}^{N_{\text{tem}}} \} $
        \Require Newest frame feature $e$
        \Require Set of all data points $\mathbf{X} = \{M_{\text{tem}}^1, M_{\text{tem}}^2, \dots, M_{\text{tem}}^{N_{\text{tem}}}, e\}$
        \Require Maximum number of iterations $T$
        \Require Weights vector of points $\mathbf{w} = \{w_1, w_2, \dots, w_{N_{\text{tem}}}, 1\}$
        \Procedure{Weighted K-means}{$\mathbf{X}, k, T, \mathbf{w}$}
            \State Number of clusters $k \gets N_{\text{tem}}$
            \State Initialize $t \gets 0$
            \State Randomly initialize cluster centroids $\mathbf{C}=\{\mathbf{c}_1, \mathbf{c}_2, \dots, \mathbf{c}_k\}$ from the data points $\mathbf{X}$
            \State Initialize previous cluster assignment $P_j \gets \{ \}$
            \State Initialize current cluster assignment $S_j \gets \{ \}$
            
            \While{$t < T$}
                \For{ $\mathbf{x}_i \in \mathbf{X}$}
                    \State $j \gets \text{argmin}_{j} \lVert \mathbf{x}_i - \mathbf{c}_j {\rVert}^2$
                    \State append ($S_j, \mathbf{x}_i$)
                \EndFor
                \If{$S == P$}
                    \State \textbf{break}
                \EndIf
                \For{ $j = 1, 2, \dots, k$}
                    \State $\mathbf{c}_j \gets \frac{\sum_{\mathbf{x}_i \in S_j} w_i \cdot \mathbf{x}_i}{\sum_{\mathbf{x}_i \in S_j} w_i}$
                \EndFor
                \State $\mathbf{w} \gets$ UpdateWeights$(S)$ \Comment{Update the weights vector based on the current cluster assignment}

                \State $P \gets S$
                \State Clear $S$
                \State $t \gets t + 1$
            \EndWhile
            $\mathbf{M_{\text{tem}}} \gets \mathbf{C} $
            \State \Return $\mathbf{M_{\text{tem}}}, \mathbf{w}$
        \EndProcedure
    \end{algorithmic}
\end{algorithm}

\begin{algorithm}
    \caption{Semantic Attention}
    \label{alg:sa}
    \begin{algorithmic}[1] 
        \Require Current abstract memory $\mathbf{M_{\text{abs}}} = \{M_{\text{abs}}^1, M_{\text{abs}}^2, \dots, M_{\text{abs}}^{N_{\text{abs}}} \} $
        \Require Newest frame features $\mathbf{e}$
        \Require Memory decay factor $\alpha \in (0, 1)$ 
        \Procedure{Semantic Attention}{$\mathbf{M_{\text{abs}}}, e, \alpha$}       
            \State $K \gets f_{\text{k\_proj}}(\mathbf{e})$
            \State $Q \gets f_{\text{q\_proj}}(M_{\text{abs}})$
            \State $W \gets Q K^T $
            \State $W \gets \text{Softmax}(W, \text{dim}=1)$
            \State $\mathbf{M_{\text{abs}}} \gets (1-\alpha) \mathbf{M_{\text{abs}}} + W \mathbf{e} $
            \State \Return $\mathbf{M_{\text{abs}}}$
        \EndProcedure
    \end{algorithmic}
\end{algorithm}

For abstract memory, we design a learning-based Semantic Attention model for information integration and selective forgetting. \Cref{alg:sa} describes the detailed forward procedure of Semantic Attention model.
In order to update abstract memory $ M_{\text{abs}} \in \mathbb{R}^{N_{\text{abs}}\times P_{\text{abs}}^2}$ with newest features $\mathbf{e} \in \mathbb{R}^{n\times P_{\text{abs}}^2}$($n$ is 1 by default), we first calculated the attention weight between newest features and current abstract memory. Then a softmax layer is applied to normalize the contribution of new features. Finally, the abstract memory is updated by a momentum updating mechanism with decay factor $\alpha$.

\begin{algorithm}
    \caption{Key Feature Retrieval}
    \label{alg:ret}
    \begin{algorithmic}[1] 
        \Require Current feature buffer $\mathbf{M_{\text{buff}}} = \{M_{\text{buff}}^1, M_{\text{buff}}^2, \dots \} $
        \Require Current temporal memory $\mathbf{M_{\text{tem}}} = \{M_{\text{tem}}^1, M_{\text{tem}}^2, \dots, M_{\text{tem}}^{N_{\text{tem}}} \} $
        \Require Weights vector of points $\mathbf{w} = \{w_1, w_2, \dots, w_{N_{\text{tem}}}\}$
        \Procedure{Key Feature Retrieval}{$\mathbf{M_{\text{buff}}}, \mathbf{M_{\text{tem}}}, \mathbf{w}, N_{ret}$}       
            \State $k \gets  N_{ret}$
            \State $j_1, j_2, \dots, j_k \gets \text{top-k}_{j}~w_j $
            \State $\mathbf{M_{\text{ret}}} \gets \{ \}$
            \For{ $z = 1, 2, \dots, k$}
                \State $e_{\text{key}} \gets \text{min\_item} {\lVert g_c(e_{\text{key}},P_{\text{spa}})-M_{\text{tem}}^{j_z} \rVert}^2$ for $e_{\text{key}} \in M_{\text{buff}}$ 
                \State append ($\mathbf{M_{\text{ret}}}, e_{\text{key}}$)
            \EndFor
            \State \Return $\mathbf{M_{\text{ret}}}$
        \EndProcedure
    \end{algorithmic}
\end{algorithm}

For retrieved memory, we use a key feature retrieval \Cref{alg:ret} to calculate the current retrieved memory $M_{\text{ret}} \in \mathbb{R}^{N_{\text{ret}} \times P_{\text{spa}}^2}$. Because retrieved memory and spatial memory are both renewed from the feature buffer $M_{\text{buff}}$, we set their spatial sizes to the same. Here $w_j$ is equal to the size of $j$-th cluster, i.e., the number of tokens in this cluster. Therefore, we choose the centroids of the top-k large clusters as pivots. The features nearest to these centroids are considered as key features, which are added to the retrieved memory.

\section{Training details}

\begin{table}[H]
    \caption{Training settings of Flash-VStream}
    \label{tab:train_setting}
    \centering
    \begin{tabular}{lcc}
        \toprule
        Settings& Stage-1& Stage-2 \\
        \midrule
        Batch size& 256 & 128 \\ 
        Learning rate& 1e-3 & 2e-5 \\ 
        Learning schedule & \multicolumn{2}{c}{Cosine decay} \\ 
        Warmup ratio & \multicolumn{2}{c}{0.03} \\ 
        Weight decay & \multicolumn{2}{c}{0} \\ 
        Epoch& \multicolumn{2}{c}{1} \\ 
        Optimizer & \multicolumn{2}{c}{AdamW} \\ 
        DeepSpeed stage& 0 & 1 \\ 
        Visual encoder & \multicolumn{2}{c}{Freeze} \\ 
        Semantic attention& \multicolumn{2}{c}{Open} \\ 
        Projector & \multicolumn{2}{c}{Open} \\ 
        LLM& Freeze & Open \\ 
        
        \bottomrule
    \end{tabular}
\end{table}

The training procedure of Flash-VStream is similar to that of \cite{llava} \cite{llamavid}. In the modality alignment stage (stage 1), we train the Semantic attention model and the projector for one epoch. In the instruction tuning stage (Stage 2), we fine-tune the Semantic attention model, the projector and the LLM for another epoch. 
The overall training can be finished in 15 hours on 8 A100 80G GPUs (BFloat16) with extracted visual features. Detailed training settings are shown in \Cref{tab:train_setting}.

\section{VStream-QA benchmark design details}
Here we provide more details of VStream-QA online video understanding benchmark.

\subsection{Data generation pipeline in detail}
\label{sec:data_gen}

\begin{itemize}[leftmargin=*]
    \item \textbf{Video Selection.} We first select 10 videos from Ego4D dataset~\cite{grauman2022ego4d} with each video being 1 hour long,
          and 22 videos from MovieNet dataset~\cite{huang2020movienet} with each video being 30 minutes long.
          Both Ego-centric videos and movie clips are chosen to cover a wide range of content types. Refer to next subsection for details.
    \item \textbf{Dense Captioning.} We use GPT-4V~\cite{OpenAI2023GPT4TR} to generate dense captions for each video clip.
          Long videos are divided into pieces of 30 seconds, and 8 frames are sparsely sampled from each piece as input to GPT-4V.
          Each output caption describes the content of the 30-second video piece, and marked with a specific timestamp.
    \item \textbf{Summary Generation.} We use GPT-4 to deduplicate and summarize the dense captions generated by GPT-4V.
          The summary is designed to be a concise description scene-level clip, typically originated from multiple dense captions that correspond to
          several minutes of video content. Timestamps are carefully kept throughout the summarization process.
    \item \textbf{Question-Answer Generation.} We use GPT-4 to generate 5 types of QA pair based on the scene summary.
          Each QA is generated from a single or several consecutive scene summaries,
          to ensure that the QA is only related to the visual information before the timestamp.
    \item \textbf{Human Filtering.} Volunteers are invited to judge the relevance of the generated QA pairs to the video content.
          The following types of QA pairs are carefully filtered out:
          i) questions are irrelevant with the video or ambiguous,
          ii) questions require additional knowledge beyond the video,
          iii) questions are able to answered without the video,
          iv) answers are wrong or ambiguous.
          repetitive.

\end{itemize}

\subsection{Variety of video content}

Besides the variety of question types, VStream-QA benchmark
also involves various type of video content.
\begin{itemize}[leftmargin=*]
    \item \textbf{VStream-QA-Ego} video topics: ['cooking', 'playing-card', 'writing', 'home-maintenance', 'sightseeing', 'reading'].
    \item \textbf{VStream-QA-Movie} movie genres: ["Action", "Adventure", "Sci-Fi", "Crime", "Drama", "Thriller", "War", "Mystery", "Comedy", "Fantasy", "History", "Biography", "Horror"].
\end{itemize}

\section{Limitations}

\subsection{Representativeness of VStream-QA benchmark}
Although the proposed VStream-QA is the first benchmark that aims to simulate real-world video streaming scenarios, it still falls short in fully representing the scenario of comprehending \textit{infinitely long video streams} in the real world.
Besides, the proposed approach only involves the coarse-grained understanding task, \textit{i.e.}, QA. In the real world, video streams encompass more complex comprehension tasks. It is our aspiration that the Flash-VStream could inspire related research in this field.

\subsection{GPT-3.5-based evaluation metric}
In the proposed VStream-QA benchmark and many other video question answering benchmarks, GPT-3.5 based evaluation is adopted as the preferred metric. However, we notice that there is always a discrepancy between the distribution of GPT accuracy and GPT score. Specifically, for answers classified as ``no'', many of them are assigned with a high score like ``4'' or ``5'', also discussed by \cite{moviechat}. This abnormal phenomenon reduces the credibility of this ``$0 \sim 5$ score'' metric in GPT-3.5-based MLLM evaluation.

\section{Broader Impacts}
Real-time understanding models for long video streams may lead to potential negative societal impacts, including but not limited to unauthorized surveillance or privacy-infringing tracking. 
However, we firmly believe that the task itself is neutral with positive applications, such as health monitoring and emergency response.


\end{document}